\documentclass[10pt,twocolumn,letterpaper]{article}

\usepackage{cvpr}              
\definecolor{cvprblue}{rgb}{0.21,0.49,0.74}

\usepackage[pagebackref,breaklinks,colorlinks,allcolors=cvprblue]{hyperref}
\usepackage{xcolor}   
\usepackage{xspace}
\usepackage{tikz}
\usepackage{caption} 
\usepackage[table]{xcolor}

\def\paperID{799} 
\def\confName{CVPR}
\def\confYear{2026}

\title{\vspace{-13pt}\cotname: Enhancing Granularity and Generalization in Image Editing \\[-20pt]} 

\author{
    Shiyi Zhang\textsuperscript{1,2,*}, 
    Yiji Cheng\textsuperscript{2,*}, 
    Tiankai Hang\textsuperscript{2,*}, 
    Zijin Yin\textsuperscript{2}, 
    Runze He\textsuperscript{2}, 
    Yu Xu\textsuperscript{2}, \\
    Wenxun Dai\textsuperscript{1,2}, 
    Yunlong Lin\textsuperscript{2}, 
    Chunyu Wang\textsuperscript{2}, 
    Qinglin Lu\textsuperscript{2}, 
    Yansong Tang\textsuperscript{1,$\dagger$}
    \\ \vspace{5pt}
    \textsuperscript{1} Shenzhen International Graduate School, Tsinghua University
    \textsuperscript{2} Hunyuan, Tencent \\
    {\tt \small \{sy-zhang23@mails., tang.yansong@sz.\}tsinghua.edu.cn}
}

\newcommand{\cotname}{Meta-CoT\xspace}
\newcommand{\rewardname}{CoT-Editing Consistency Reward\xspace}
\newcommand{\redword}[1]{\textcolor{red}{#1}}
\newenvironment{zeroparfillskip}{%
  \parfillskip=0pt\relax
}{%
  \par 
}

\begin{document}

\makeatletter
\let\@oldmaketitle\@maketitle
\renewcommand{\@maketitle}{\@oldmaketitle
\begin{minipage}{\textwidth}
\vspace{-10mm}
\centering
\includegraphics[width=1\linewidth]{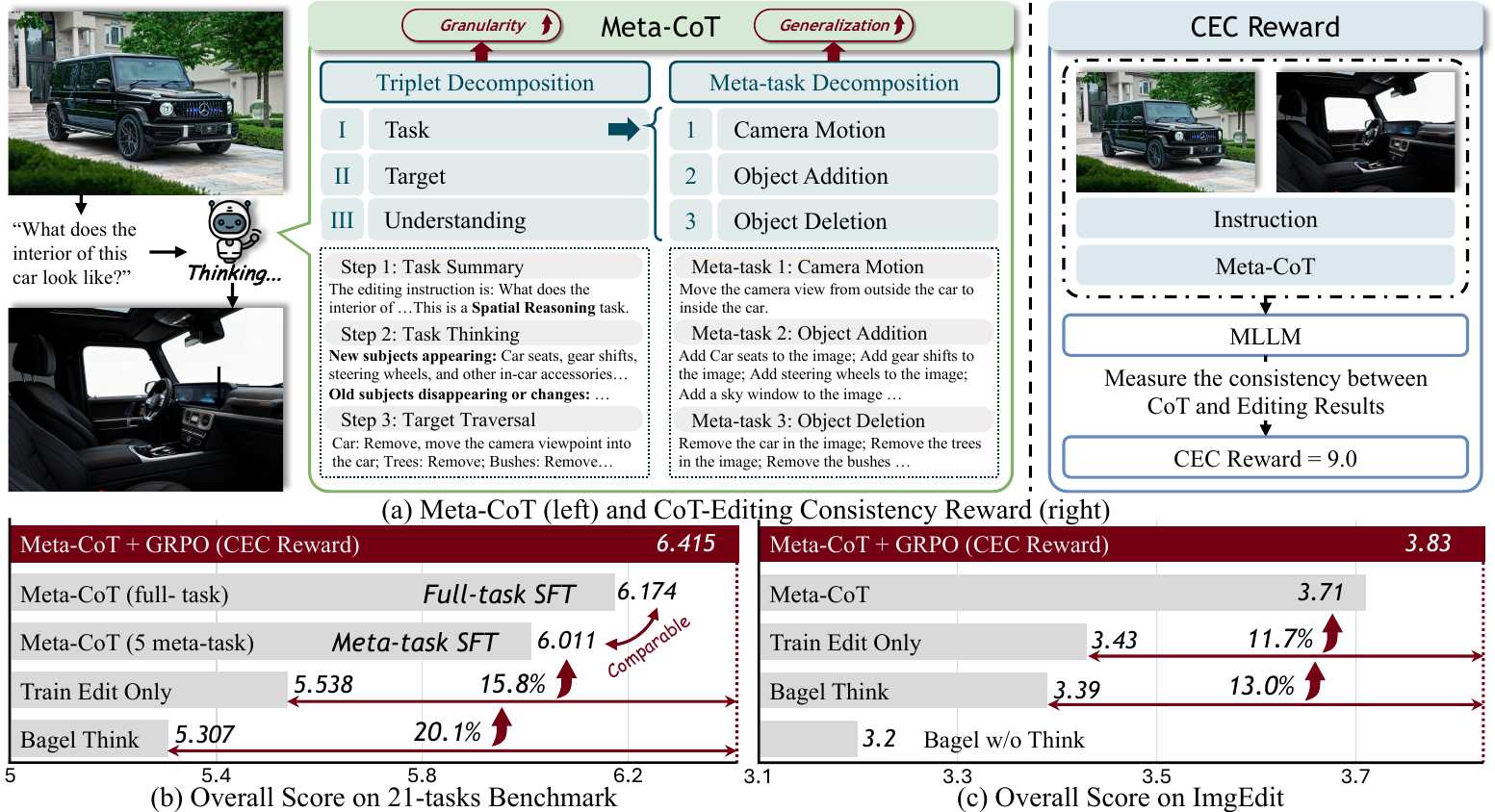}
    \vspace{-6.5mm}
    \captionsetup{hypcap=false}
    \captionof{figure}{(a) We propose \cotname with a hierarchical decomposition design and the CoT–Editing Consistency Reward.
On (b) 21-task Editing Benchmark and (c) ImgEdit, our method shows:
\textbf{(1) Stronger instruction-based editing.} It achieves 15.8\% and 11.7\% gains over the Train-Edit-Only baseline trained on the same parameters and editing data (without Meta-CoT).
\textbf{(2) Generalization to unseen tasks.} As shown in (b), meta-task training enables comparable performance to full-task training while training only on a small subset of meta-tasks.}

    \label{fig:teaser}
    \vspace{+2mm}
\end{minipage}}
\vspace{-12pt}
\makeatother

\maketitle
\renewcommand{\thefootnote}{\fnsymbol{footnote}}
\footnotetext[1]{Equal contribution.}
\footnotetext[2]{Corresponding author.}
\renewcommand{\thefootnote}{\arabic{footnote}}
\begin{abstract}
Unified multi-modal understanding/generative models have shown improved image editing performance by incorporating fine-grained understanding into their Chain-of-Thought (CoT) process. 
However, a critical question remains underexplored: what forms of CoT and training strategy can jointly enhance both the understanding granularity and generalization? To address this, we propose Meta-CoT, a paradigm that performs a two-level decomposition of any single-image editing operation with two key properties: 
\textbf{(1) Decomposability.} We observe that any editing intention can be represented as a triplet — (task, target, required understanding ability). Inspired by this, Meta-CoT decomposes both the editing task and the target, generating task-specific CoT and traversing editing operations on all targets. This decomposition enhances the model's understanding granularity of editing operations and guides it to learn each element of the triplet during training, substantially improving the editing capability.
\textbf{(2) Generalizability.} In the second decomposition level, we further break down editing tasks into five fundamental meta-tasks. We find that training on these five meta-tasks, together with the other two elements of the triplet, is sufficient to achieve strong generalization across diverse, unseen editing tasks.
To further align the model's editing behavior with its CoT reasoning, we introduce the \textbf{\rewardname}, which encourages more accurate and effective utilization of CoT information during editing. Experiments demonstrate that our method achieves an overall 15.8\% improvement across 21 editing tasks, and generalizes effectively to unseen editing tasks when trained on only a small set of meta-tasks. Our code, benchmark, and model are released at \href{https://shiyi-zh0408.github.io/projectpages/Meta-CoT/ }{here}.
\end{abstract}    
\begin{figure*}[t]
    \centering
    \includegraphics[width=\linewidth]{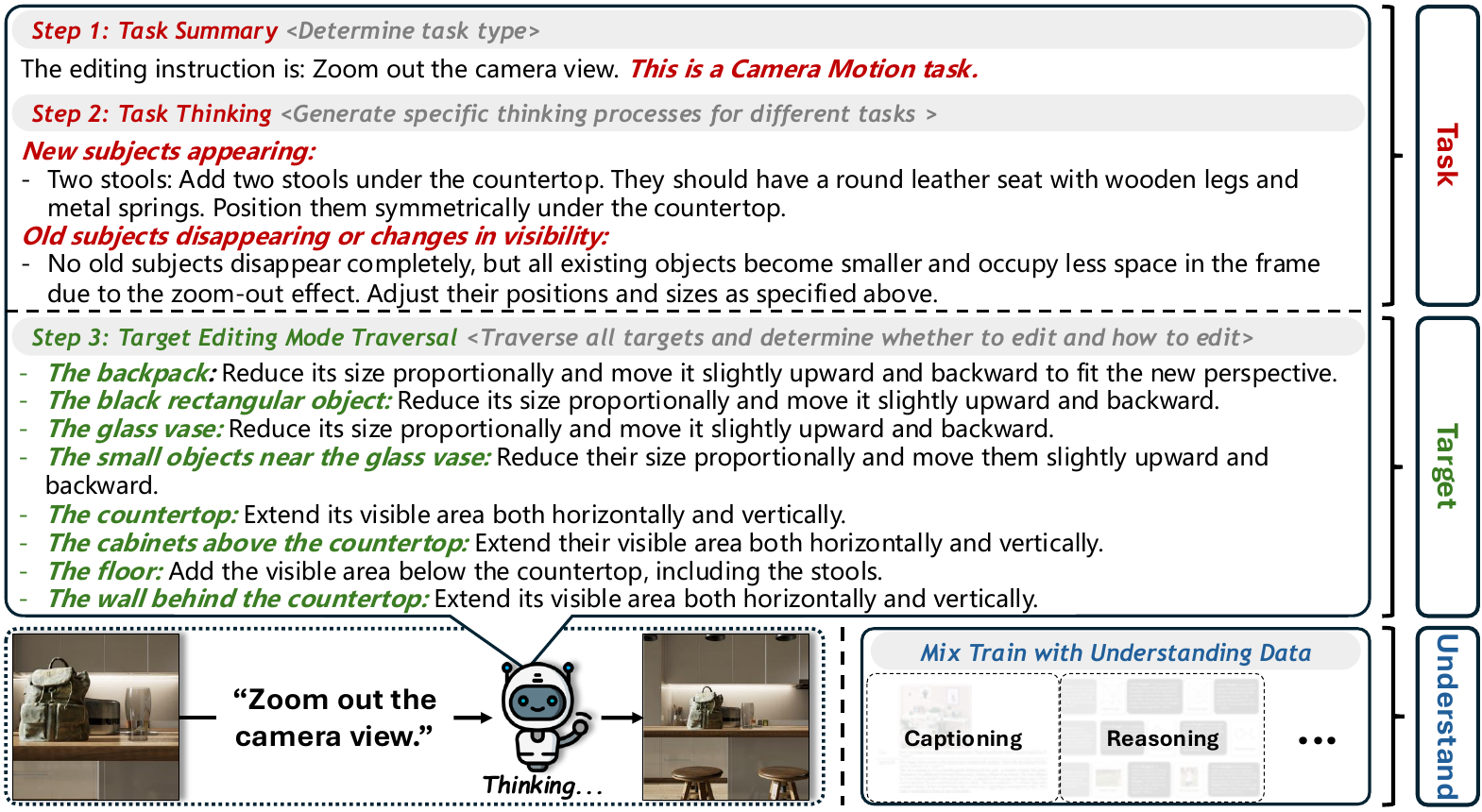}
    \vspace{-20pt}
    \caption{\textbf{Overview of Triplet Decomposition.}
Triplet Decomposition enables fine-grained reasoning over both the task and the target through three steps: (1) Task Summary, (2) Task Thinking, and (3) Target-wise Editing Traversal. This decomposition allows the model to optimize reasoning and editing along two elements: task and target. During training, we jointly incorporate diverse visual understanding tasks to capture all components of the (task, target, understanding capability) triplet, as illustrated in the bottom-right.}
    \label{fig:overview}
    \vspace{-10pt}
\end{figure*}
\section{Introduction}
\label{sec:intro}



\begin{zeroparfillskip}
Recent studies show that Chain-of-Thought (CoT) can effectively enhance editing performance in unified generation/understanding models~\cite{fang2025got,deng2025emerging}. By explicitly reasoning the editing process, the model activates its understanding ability and produces more accurate edits. We argue that an effective image editing CoT paradigm should possess two properties: \textbf{(1) Stimulation of the model's understanding ability.} \textbf{(2) Generalization across diverse editing tasks.}

Existing works mainly focus on the first property, such as introducing spatial localization cues or other explicit understanding information into the CoT~\cite{fang2025got}. However, such approaches often exhibit limited generality since CoTs tailored to specific understanding forms may not adapt well to broader editing tasks. For example, localization-based CoT performs poorly on tasks like style transfer or viewpoint transformation. Therefore, this paper aims to investigate a critical question: what forms of CoT and training strategy can simultaneously enhance understanding granularity and generalization capability during image editing?
To address this, we first observe that any single-image editing operation can be defined by a triplet: (Task, Target, Required Understanding Ability).
For example, the editing instruction ``Change the number of puppies to three" involves the task type ``Quantity Modification", editing targets ``puppies", and requires the model to possess understanding abilities in localization and counting.
Building on this, as shown in Figure \ref{fig:overview}, we propose the first decomposition level in the Meta-CoT, Triplet Decomposition, which guides fine-grained reasoning over both the editing task and the editing target. This paradigm comprises three steps: (1) Task Summary, (2) Task Thinking, and (3) Target-wise Editing Traversal.
This design offers strong decomposability, enabling the model to optimize over task and target, thereby not only learning to comprehend diverse editing operations but also mastering how to apply distinct editing strategies to different entities.
To capture the third element, required understanding ability, of the triplet, we incorporate data from a variety of visual understanding tasks during training, ensuring the model fully learns across all three elements.

\begin{table*}[t]
\begin{center} 
\fontsize{8}{14}\selectfont \renewcommand{\arraystretch}{0.8}
\caption{Any single-image editing operation can be represented as a triplet of Task (first column), Editing Target (third column), and Required Understanding Capability (fourth column). Each operation can be further decomposed into combinations of Meta-Tasks. We summarize five generic meta-tasks (top) and list their possible compositional combinations for different editing tasks (second column).}
\vspace{-5pt}
\label{tab:triplet}
\fontsize{8}{14}\selectfont \renewcommand{\arraystretch}{0.67}
\setlength{\tabcolsep}{2mm}
\begin{tabular}{l|c|c|c}
\toprule
Task & Meta-Task & Target & Understanding \\ 
\midrule
Addition & Addition & Any & Localization \\ 
Deletion & Deletion & Any & Localization \\ 
Replacement & Replacement & Any & Localization \\ 
Camera Motion & Camera Motion & Any & Spatial Understanding \\ 
Position Change & Position Change & Any & Spatial Understanding \\ \hline
Style Transfer & Replacement & Style & Style Understanding \\ 
Tone Adjustment & Replacement & Tone / Lighting & Tone Understanding \\ 
Text Editing & Addition, Deletion, Replacement & Text & Image-text Understanding / OCR \\ 
Shape Modification & Replacement & Shape & Shape Understanding \\ 
Structural Change & Addition, Deletion, Replacement, Camera Movement & Artificial Structure & Structural Understanding \\ 
Color Change & Replacement & Color & Color Understanding \\ 
Quantity Change & Addition, Deletion & Any & Counting \\ 
Specified Quantity Change & Any & Any & Counting, Localization \\ 
Human Attribute Editing & Addition, Deletion, Replacement & Human & Human Attribute Understanding \\ 
Material Change & Replacement & Material & Material Understanding \\ 
Motion Change & Replacement & Motion & Motion Understanding \\ 
Logical Reasoning & Addition, Deletion, Replacement & Shape, Symbol, Text & Symbolic / Mathematical Reasoning \\ 
Causal Reasoning & Addition, Deletion, Replacement & Any & Causal Reasoning \\ 
Spatial Composition & Addition, Deletion, Position Change, Camera Movement & Any & Spatial Reasoning \\ 
Temporal Reasoning & Addition, Deletion, Replacement & Any & Temporal Reasoning \\ 
Multi-Instruction Editing & Any & Any & Multi-instruction Understanding \\ 
\bottomrule
\end{tabular}
\end{center}
\vspace{-20pt}
\end{table*}


Furthermore, to achieve generalization, we analyze single-image editing tasks and identify a set of primitive, universal operations, termed ``meta-tasks".
As shown in Table \ref{tab:triplet} and Figure \ref{fig:metatask}, analogous to bases in a vector space, these meta-tasks (e.g., add, remove, replace) form a minimal set capable of spanning the entire editing operation space.
We then propose the second decomposition level in Meta-CoT: Meta-task Decomposition. Specifically, we redefine the triplet element ``Task" as one or more meta-tasks, and adapt the CoT paradigm by replacing ``Task Summary" with ``Meta-task Summary". For example, ``Quantity Change" corresponds to the meta-tasks add and remove. The model learns to decompose instructions into simple meta-tasks, forming a reasoning chain of fundamental operations.
This paradigm yields strong generalizability: by training on only a small number of meta-tasks (as few as five), the model can generalize to all other editing tasks via compositional reasoning.
Thus, during training, we no longer need to cover every task individually; all remaining tasks can be expressed as combinations of these meta-tasks.
We also incorporate diverse visual understanding tasks into the training data to ensure mastery of all triplet elements.

\begin{figure}[t]
    \centering
    \includegraphics[width=\linewidth]{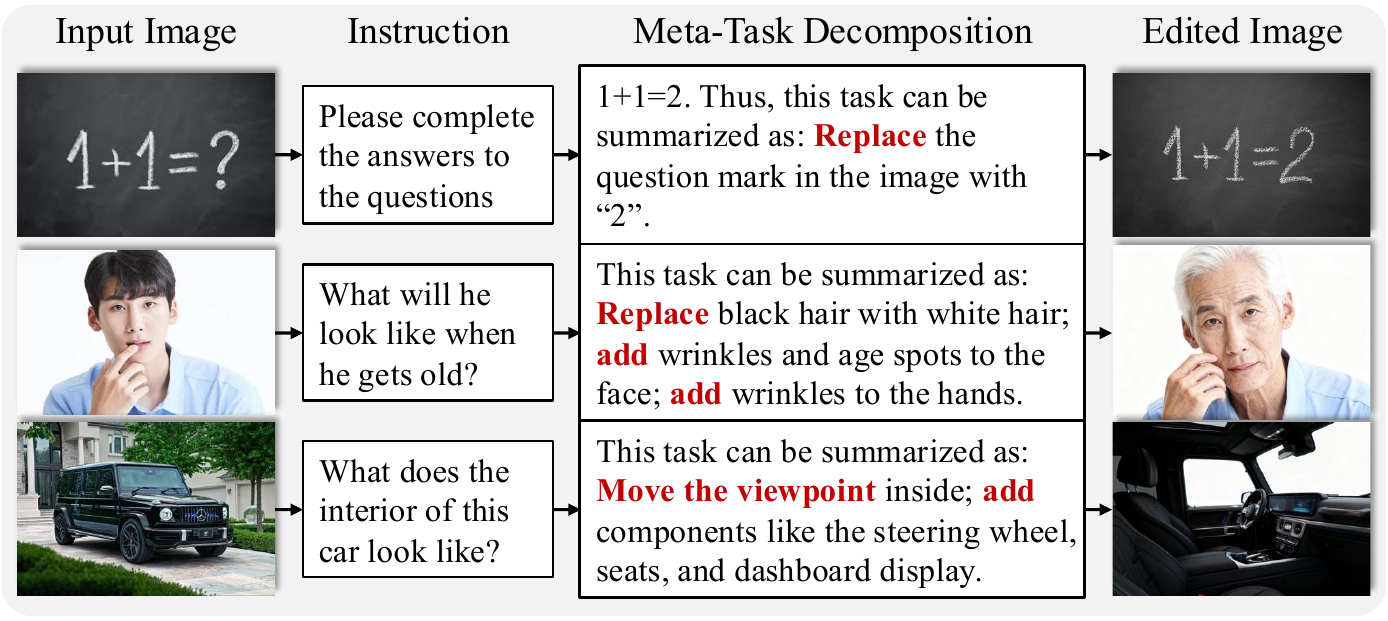}
    \vspace{-20pt}
    \caption{Examples of editing intent decomposed into meta-tasks.}
    \label{fig:metatask}
    \vspace{-20pt}
\end{figure}

We further observe the inconsistency between the CoT reasoning and the final edited result in existing models. To address this, we propose the CoT–Editing Consistency Reward.
Specifically, we employ a VLM (e.g., Qwen2.5-VL~\cite{qwen2.5-vl}) to assess the consistency between the CoT and the edited image from both the task and object perspectives, and provide corresponding rewards.
Using this reward, we perform Flow-GRPO~\cite{liu2025flow} and effectively improve the alignment between the CoT reasoning and editing outcomes.

To validate our approach, we construct a benchmark covering 21 distinct editing tasks, including Gedit-Bench~\cite{liu2025step1x}, RiseBench~\cite{rise}, ComplexEdit~\cite{yang2025texttt}, and five additional benchmarks we developed to fill missing task types. Compared to existing benchmarks, our benchmark offers broader task coverage, enabling a more thorough evaluation.
The benchmark data is distributionally independent from training data, ensuring fair evaluation.
We also conduct experiments on ImgEdit~\cite{ye2025imgedit} to demonstrate the effectiveness of our method.
We conduct our experiments based on Bagel~\cite{deng2025emerging}, a mainstream open-source unified model that supports CoT–based editing, making it well-suited for our comparative study.
After applying Meta-CoT via SFT and the CoT–Editing Consistency Reward via GRPO optimization, our method achieves a 15.7\% overall performance improvement across all 21 tasks compared to the base model. Moreover, experiments demonstrate that training on only a few meta-tasks is sufficient to achieve strong generalization to other editing tasks, yielding performance comparable to that achieved by training on the full set of task types.


The contributions of this paper can be summarized as:
\begin{itemize}
    \item We propose Triplet Decomposition, which breaks down editing tasks into task, target, and understanding ability, enhancing the granularity during the reasoning process.
    \item We introduce Meta-task Decomposition, which reduces tasks to fundamental meta-tasks, enabling strong generalization through the meta-task-based training strategy.
    \item We design a \rewardname to mitigate mismatches between CoT reasoning and editing results.
    \item We construct a comprehensive image editing benchmark which covers a broader range of editing task types and empirically validate the effectiveness of our approach.
\end{itemize}
\end{zeroparfillskip}

\section{Related Work}
\label{sec:relatedwork}

\subsection{CoT in Understanding/Generation Models} 
\begin{zeroparfillskip}
Chain-of-Thought (CoT) reasoning, originally introduced to elicit step-by-step logical inference in large language models, has recently been extended to vision-language models (VLMs) to enhance multimodal reasoning~\cite{zhang2023multimodal, wei2022chain, wang2022self, huang2022large, lu2022learn}. Early efforts integrated visual information into textual CoT~\cite{zhang2023multimodal, lu2022learn}, while subsequent work developed native multimodal CoT by interleaving textual reasoning with visual tokens~\cite{gupta2023visual, wu2024v, hu2024visual, shao2024visual, fu2025refocus, li2025imagine}. 
CoT has also been applied to image generation~\cite{huang2025interleaving, han2025controlthinker,duan2025got,jiang2025t2i} and editing~\cite{deng2025emerging,fang2025got}. In image editing, several works have incorporated CoT to improve instruction following and spatial reasoning. For instance, Bagel~\cite{deng2025emerging} and GoT~\cite{fang2025got} generate intermediate reasoning steps before editing images, demonstrating improved consistency between visual understanding and generation. 
However, current CoT paradigms for editing are either too generic to stimulate the model's understanding capacity~\cite{deng2025emerging} or too specialized to adapt across diverse tasks~\cite{fang2025got}, motivating us to develop a paradigm that enhances both the understanding granularity and generalization in image editing.

\end{zeroparfillskip}

\subsection{Unified Understanding and Generation Models}
\begin{zeroparfillskip}
Recent research has increasingly focused on unifying multimodal understanding~\cite{Qwen-VL, qwen2.5-vl, zhang2023logo, zhang2024narrative} and generation~\cite{peebles2023scalable, zhang2025flexiact, zhu2025kv, li2025tooncomposer, li2025nvcomposer, ma2024followpose, ma2026fastvmt, xu2024headrouter} within single models for joint understanding and synthesis~\cite{sun2024emu, chameleon, wu2025janus, xie2024show, chen2025blip3, liao2025mogao, xie2025show, wu2025omnigen2, deng2025emerging, huang2025ming, xu2026tag, dai2026chatumm, he2026re, lin2025jarvisevo, lin2025jarvisart, zhuang2024colorflow, yang2024direct, yang2023uni, he2025freeedit}. GPT‑4o~\cite{openai2025chatgpt4o} exemplifies this by fusing visual analysis and generation, outperforming earlier unified models. Current unified models can be organized into three categories: (1) Autoregressive approaches that apply next-token prediction to jointly generate textual and visual tokens~\cite{wu2025janus, januspro2025, lu2024unified, qu2024tokenflow, chameleon, emu3}; (2) Additional Diffusion frameworks that couple a pre-trained LLM backbone with an external diffusion module, where the language model derives semantic conditions to guide the diffusion process~\cite{dream-llm, wu2024next, pan2025transfer, tong2024metamorph}; and (3) Unified Integrated Transformer models that natively combine both LLM and diffusion mechanisms within a shared transformer architecture~\cite{deng2025emerging,liang2024mixture,janusflow2024,shi2024llamafusion,transfusion}. We build our method upon Bagel~\cite{deng2025emerging}, a unified multimodal transformer with intrinsic CoT-based editing capabilities, making it a suitable foundation for CoT paradigms exploration in image editing.
\end{zeroparfillskip}

\section{Method}

\subsection{Theoretical Definition}
\begin{zeroparfillskip}
\noindent\textbf{Triplet Decomposition.}
Let $\mathcal{T}$ denote the original CoT space, and let $T_{\text{1}}$, $T_{\text{2}}$, and $T_{\text{3}}$ denote the task, target, and required understanding capability. Then the triplet space is $\mathcal{S}_{triplet} = T_{\text{1}} \times T_{\text{2}} \times T_{\text{3}}$. We further let $H = \log|\text{space}|$ denote space complexity (i.e., entropy). We prove:
$ H(T_{\text{1}}, T_{\text{2}}, T_{\text{3}}) = \log|\mathcal{S}_{triplet}| < \log|\mathcal{T}| = H(\mathcal{T})$, indicating triplet decomposition lowers the editing complexity. Next, we use the mutual information per entropy $G = \frac{I(T; X_{\text{tgt}})}{H(T)}$~\cite{theil1970estimation} to denote the understanding granularity in CoT, where $X_{tgt}$ is the target image. We prove: $G(T_{\text{1}}, T_{\text{2}}, T_{\text{3}})>G(T)$
, indicating Meta-CoT achieves higher understanding granularity than classical CoT (noted as $T$). See supplementary material for detailed derivation.
\end{zeroparfillskip}

\noindent\textbf{Meta-task Decomposition.}
$\mathcal{B}=\{t_1,\ldots,t_n\}$ is a set of meta-tasks.
We say $\mathcal{B}$ forms a basis of the task space $\mathcal{T}$ if
\vspace{-6pt}
\[
\forall T\in\mathcal{T},\;
\exists\, t_{i_1},\ldots,t_{i_k}\in\mathcal{B}
\quad\text{s.t.}\quad
T=t_{i_1}\circ t_{i_2}\circ\cdots\circ t_{i_k}.
\]
\vspace{-18pt}

\subsection{Triplet Decomposition}
\label{subsec:taskcot}
As shown in Table \ref{tab:triplet}, any single-image editing operation can be decomposed into a triplet comprising three fundamental elements: the editing task, the editing target, and the understanding capability required to accomplish the operation. Building on this insight, we propose the first decomposition level, Triplet Decomposition (illustrated in Figure \ref{fig:overview}), in \cotname. This paradigm decomposes an editing instruction into task and target, guiding the model to explicitly learn to understand different editing tasks and master editing methods for various targets during training. To accommodate the diverse understanding capabilities required for different editing tasks, we incorporate a diverse set of visual understanding tasks during training, ensuring the comprehensive mastery of all three elements of the defined triplet.

\begin{zeroparfillskip}
As shown in Figure \ref{fig:overview}, our Triplet Decomposition unfolds in three steps. \textbf{(1) Task Summary.} The model infers the task type inductively from the instruction. \textbf{(2) Task Thinking.} The model generates a task-specific reasoning process based on the task type. For instance, for style transfer, it analyzes the visual attributes of the target style; for camera motion, it identifies object appearance or disappearance; and for logical reasoning-based editing, it deduces the implicit operations suggested by the instruction (see the supplementary materials for more examples). \textbf{(3) Target Editing Mode Traversal.} The model traverses all targets in the image and reasons about whether and how each should be edited. This step ensures spatial and semantic consistency and provides fine-grained, interpretable editing guidance.
\end{zeroparfillskip}

\subsection{Meta-task Decomposition}
\label{subsec:metatasktraining}
\begin{zeroparfillskip}
Furthermore, as shown in Table \ref{tab:triplet}, we identify a set of generic operations within single-image editing, referred to as ``meta-tasks". Meta-tasks serve as a set of bases in the single-image editing operation space, capable of combining and generalizing to produce various complex editing operations. In the ideal case, training on these meta-tasks enables the model to handle other, more complex editing tasks.
\end{zeroparfillskip}

\begin{zeroparfillskip}
Building on this insight, we propose the second decomposition level, Meta-task Decomposition, in \cotname. In practice, we define five distinct meta-tasks (listed in Table \ref{tab:triplet}). Accordingly, our triplet evolves into (meta-task, target, required understanding capability). Next, we replace the first step of \cotname, Task Summary, with Meta-task Summary, decomposing the instruction into a combination of basic meta-tasks. For example, the style transfer task can be decomposed into a ``replacement" operation on the style attribute of the editing target. During training, as noted in Section \ref{subsec:taskcot}, we supervise all triplet elements by training on data of the five meta-tasks and incorporating diverse visual understanding tasks to achieve comprehensive mastery.
\end{zeroparfillskip}

\subsection{CoT-Editing Consistency Reward}
\label{subsec:consistencyreward}
In our experiments, we observed that in certain editing scenarios, particularly when the editing instruction does not explicitly specify the operation, the model may fail to follow the reasoning outlined in the CoT, even if the correct editing operation is inferred. This misalignment between CoT reasoning and execution often results in incorrect editing.

To address this, we introduce the CoT–Editing Consistency (CEC) Reward.
Specifically, we design a consistency metric from both the task and target perspectives, where a VLM (Qwen2.5-VL~\cite{qwen2.5-vl}) evaluates whether the generated edit aligns with the CoT reasoning in terms of both operation and target, producing a score from 0 to 10.
Before training, we conduct a correlation validation for the CEC Reward. Specifically, we use the Meta-CoT SFT model to generate 500 editing samples with CoT. Four annotators then score the CoT–editing consistency. For samples with scores ranging $<3$, we average all scores. For those with a range $\ge3$, we average the three closest scores. We then iteratively adjust the VLM's initial prompt, computing the Pearson correlation $r$ and mean absolute error $\epsilon_{\text{MAE}}$ against human annotations on the 500 samples, until $r \ge 0.8$ and $\epsilon_{\text{MAE}} \le 2.5$~\cite{lee2024prometheus,yasunaga276482127multimodal,ku2023viescore}. We provide the details for the CEC Reward in the supplementary material.
%

\begin{figure}[t]
    \centering
    \includegraphics[width=\linewidth]{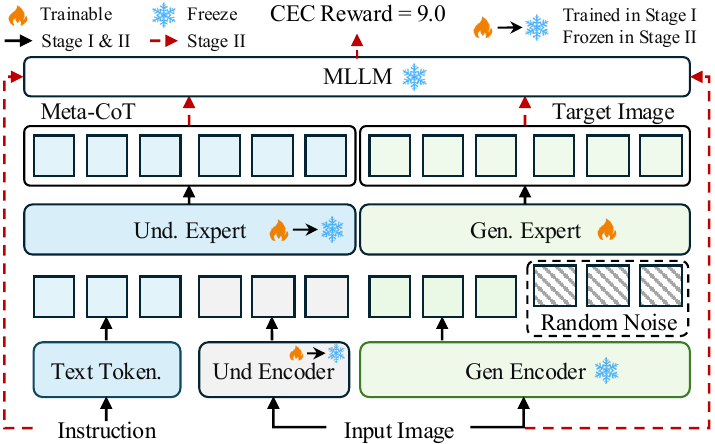}
    \vspace{-20pt}
    \caption{\textbf{Training pipeline} (self attention omitted). (1) Stage 1: SFT on both reasoning and editing. (2) Stage 2: GRPO on editing.}
    \label{fig:framework}
    \vspace{-15pt}
\end{figure}

\begin{zeroparfillskip}
We adopt Flow-GRPO~\cite{liu2025flow,shao2024deepseekmath} to optimize the model with the CEC Reward. Since the CEC Reward measures semantic alignment, we only focus the optimization on early denoising timesteps, where semantic fidelity is most critical, and omit updates on later timesteps. Empirically, we find that reducing optimization on later timesteps also alleviates potential noise artifacts introduced by Flow-GRPO.
\end{zeroparfillskip}

\subsection{Training Pipeline}
\begin{zeroparfillskip}
As shown in Figure \ref{fig:framework}, our training includes two stages. In the SFT stage, both the understanding expert and generation expert are tuned to train CoT reasoning and image editing, with the image understanding encoder also updated. 
In the subsequent RL stage, we freeze the image understanding encoder and train only the generation expert. This is motivated by two observations: (1) after SFT, the model already achieves highly accurate CoTs that should be preserved; (2) training both modules during RL causes unstable optimization and degrades the reasoning ability learned in SFT.
\end{zeroparfillskip}

\begin{figure}[t]
    \centering
    \includegraphics[width=\linewidth]{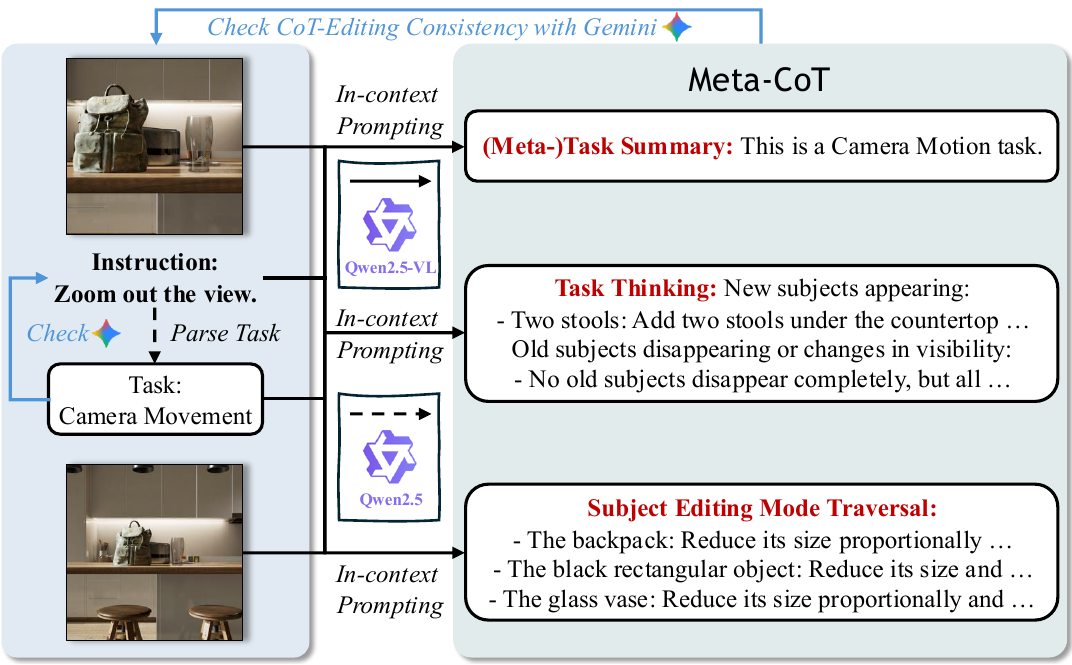}
    \vspace{-20pt}
    \caption{
    \begin{zeroparfillskip}
    \textbf{Meta-CoT Data Construction.} This pipeline processes the source image, target image, and instruction to the Meta-CoT.
    \end{zeroparfillskip}
    }
    \label{fig:data}
    \vspace{-15pt}
\end{figure}
\subsection{Meta-CoT Data Creation Pipeline}
\label{subsec:metacotdatapipeline}
\begin{zeroparfillskip}
As shown in Figure~\ref{fig:data}, for each editing sample, we first determine its editing task type with Qwen2.5~\cite{qwen2.5}, based on carefully designed task definition rules and the instruction, followed by a consistency check between the predicted task type and instruction with Gemini-2.5-Flash. Next, we input the source image, target image, instruction, and task type into Qwen2.5-VL~\cite{qwen2.5-vl}, which, guided by a carefully designed prompt, generates the (Meta-)Task Summary, Task Thinking, and Target Editing Mode Traversal. This process also includes an evaluation to verify the alignment between the generated Meta-CoT and the actual editing process.
\end{zeroparfillskip}

\section{Experiment}
\begin{table*}[t]
\centering
\fontsize{8}{14}\selectfont \renewcommand{\arraystretch}{0.67}
\caption{\textbf{Comparison of Overall Scores on the 21-task benchmark.} All metrics are evaluated using GPT-4.1. Train Editing Only denotes the setting trained with the same parameters and editing data as our method, but without Meta-CoT.}
\vspace{-10pt}
\setlength{\tabcolsep}{1.23mm}
\begin{tabular}{l|ccccccccccc}
\toprule
\textbf{Method / Task} &
Background &
Color &
Material &
Action &
\begin{tabular}[c]{@{}c@{}}Human\\ Attribute\end{tabular} &
Style &
Add &
Remove &
Replace &
Text &
Tone \\
\midrule
Bagel(w/o think) & 6.172 & 6.668 & 6.024 & 4.182 & 4.933 & 6.760 & 7.244 & 6.319 & 6.774 & \textbf{5.170} & 6.323 \\
Bagel(w think) & 6.686 & 6.469 & 5.969 & 3.997 & 3.771 & 6.156 & 7.338 & 6.272 & 6.430 & 2.124 & 6.511 \\
Train Editing Only & 6.743 & 6.537 & 6.052 & 4.155 & 4.257 & 6.571 & 7.363 & 6.296 & 6.687 & 2.651 & 6.547 \\
SFT(\cotname) & 7.173 & 7.201 & 6.493 & 4.470 & 4.760 & 6.965 & 7.640 & 7.833 & 6.931 & 3.271 & 7.236 \\
\midrule
\rowcolor{gray!20}\textbf{Meta-CoT + RL (Ours)} & \textbf{7.251} & \textbf{7.323} & \textbf{6.636} & \textbf{4.574} & \textbf{4.956} & \textbf{7.035} & \textbf{7.762} & \textbf{8.129} & \textbf{7.065} & 3.328 & \textbf{7.318} \\
\midrule
\textbf{Method / Task} &
Causal &
Logical &
\begin{tabular}[c]{@{}c@{}}Spatial\\ Reasoning\end{tabular} &
Temporal &
Camera &
Structure &
Position &
Quantity &
\begin{tabular}[c]{@{}c@{}}Specified\\ Quantity\end{tabular} &
\begin{tabular}[c]{@{}c@{}}Multi-\\ Instruction\end{tabular} &
\multicolumn{1}{|c}{\textbf{Average}} \\ 
\midrule
Bagel(w/o think) & 4.628 & 2.994 & 4.706 & 3.982 & 6.364 & 6.492 & 5.838 & 5.581 & 5.177 & 6.799 & \multicolumn{1}{|c}{5.673} \\
Bagel(w think) & 5.561 & 3.146 & 4.718 & 5.477 & 5.572 & 5.188 & 4.549 & 4.530 & 4.909 & 6.066 & \multicolumn{1}{|c}{5.307} \\
Train Editing Only & 5.710 & 3.217 & 4.683 & 5.663 & 5.964 & 5.629 & 5.177 & 4.957 & 5.093 & 6.349 & \multicolumn{1}{|c}{5.538} \\
SFT(\cotname) & 6.647 & 3.526 & 4.858 & 6.233 & 6.876 & 7.309 & 5.862 & 6.468 & 6.019 & 6.936 & \multicolumn{1}{|c}{6.224} \\
\midrule
\rowcolor{gray!20}\textbf{Meta-CoT + RL (Ours)} & \textbf{6.953} & \textbf{4.014} & \textbf{5.046} & \textbf{6.376} & \textbf{7.100} & \textbf{7.557} & \textbf{6.185} & \textbf{6.712} & \textbf{6.322} & \textbf{7.077} & \multicolumn{1}{|c}{\textbf{6.415}} \\
\bottomrule
\end{tabular}
\vspace{-10pt}
\label{tab:all_task}
\end{table*}

\begin{table*}[t]
\centering
\fontsize{8}{14}\selectfont \renewcommand{\arraystretch}{0.67}
\caption{\textbf{System comparison on ImgEdit.} All metrics are evaluated by GPT-4.1. \textit{Overall} denotes average score across all tasks.}
\vspace{-10pt}
\setlength{\tabcolsep}{2.32mm}
\begin{tabular}{l|ccccccccc|c}
\toprule
\textbf{Method} & Add & Adjust & Extract & Replace & Remove & Background & Style & Hybrid & Action & \textbf{Overall} $\uparrow$ \\
\midrule
\multicolumn{11}{c}{\textit{Closed-source Models}} \\
\midrule
FLUX.1 Kontext [Pro]~\citep{labs2025kontext} & 4.25 & 4.15 & 2.35 & 4.56 & 3.57 & 4.26 & 4.57 & 3.68 & 4.63 & 4.00 \\
GPT Image 1 [High]~\citep{gptimage} & 4.61 & 4.33 & 2.90 & 4.35 & 3.66 & 4.57 & 4.93 & 3.96 & 4.89 & 4.20 \\
\midrule
\multicolumn{11}{c}{\textit{Multi-round Editing Models}} \\
\midrule
IEAP~\cite{hu2025image} & 2.34 & 3.08 & 2.03 & 3.62 & 3.45 & 1.81 & 3.19 & 2.68 & 2.76 & 2.77 \\
\midrule
\multicolumn{11}{c}{\textit{Diffusion Only Models}} \\
\midrule
MagicBrush~\citep{zhang2023magicbrush} & 2.84 & 1.58 & 1.51 & 1.97 & 1.58 & 1.75 & 2.38 & 1.62 & 1.22 & 1.90 \\
Instruct-Pix2Pix~\citep{brooks2023instructpix2pix} & 2.45 & 1.83 & 1.44 & 2.01 & 1.50 & 1.44 & 3.55 & 1.20 & 1.46 & 1.88 \\
AnyEdit~\citep{yu2025anyedit} & 3.18 & 2.95 & 1.88 & 2.47 & 2.23 & 2.24 & 2.85 & 1.56 & 2.65 & 2.45 \\
UltraEdit~\citep{zhao2024ultraedit} & 3.44 & 2.81 & 2.13 & 2.96 & 1.45 & 2.83 & 3.76 & 1.91 & 2.98 & 2.70 \\
OmniGen~\citep{xiao2025omnigen} & 3.47 & 3.04 & 1.71 & 2.94 & 2.43 & 3.21 & 4.19 & 2.24 & 3.38 & 2.96 \\
ICEdit~\citep{zhang2025context} & 3.58 & 3.39 & 1.73 & 3.15 & 2.93 & 3.08 & 3.84 & 2.04 & 3.68 & 3.05 \\
Step1X-Edit~\citep{liu2025step1x} & 3.88 & 3.14 & 1.76 & 3.40 & 2.41 & 3.16 & 4.63 & 2.64 & 2.52 & 3.06 \\
Qwen-Image~\citep{wu2025qwen} & 4.38 & 4.16 & 3.43 & 4.66 & 4.14 & 4.38 & 4.81 & 3.82 & 4.69 & 4.27 \\
\midrule
\multicolumn{11}{c}{\textit{Unified Models}} \\
\midrule
GoT~\citep{fang2025got} & 3.61& 2.94& 1.35& 2.78& 2.57& 2.29& 3.51& 1.75& 2.66& 2.61\\
Ming-UniVision~\citep{huang2025ming} & 3.55 & 3.14 & 1.52 & 3.25 & 3.29 & 2.77 & 3.99 & 2.74 & 3.91 & 3.06 \\
BAGEL(w/o think)~\citep{deng2025emerging} & 3.56 & 3.31 & 1.70 & 3.38 & 2.62 & 3.24 & 4.49 & 2.38 & 4.17 & 3.20 \\
UniWorld-V1~\citep{lin2025uniworldv1} & 3.82 & 3.64 & 2.27 & 3.47 & 3.24 & 2.99 & 4.21 & 2.96 & 2.74 & 3.26 \\
BAGEL(think)~\citep{deng2025emerging} & 3.65 & 3.53 & 2.03 & 3.60 & 3.03 & 3.45 & 4.43 & 2.59 & 4.22 & 3.39 \\
OmniGen2~\citep{wu2025omnigen2} & 3.57 & 3.06 & 1.77 & 3.74 & 3.20 & 3.57 & 4.81 & 2.52 & 4.68 & 3.44 \\
BLIP3o-NEXT~\cite{chen2025blip3o} & \textbf{4.00} & 3.78 & 2.39 & 4.05 & 2.61 & \textbf{4.30} & 4.64 & 2.67 & 4.13 & 3.62 \\
\midrule
\rowcolor{gray!20}\textbf{Meta-CoT+RL(Ours)} & 3.87 & \textbf{3.91} & \textbf{2.40} & \textbf{4.22} & \textbf{3.74} & 3.98 & \textbf{4.80} & \textbf{3.26} & \textbf{4.33} & \textbf{3.83} \\
\rowcolor{gray!20}$\Delta$ Over Base Model & \redword{+6.0\%} & \redword{+10.8\%} & \redword{+18.2\%} & \redword{+17.2\%} & \redword{+23.4\%} & \redword{+15.4\%} & \redword{+8.4\%} & \redword{+25.9\%} & \redword{+2.6\%} & \redword{+13.0\%} \\
\bottomrule
\end{tabular}
\vspace{-18pt}
\label{tab:imgedit}
\end{table*}

\subsection{Implementation Details}
\noindent\textbf{Benchmark and Metrics.}
\begin{zeroparfillskip}
To comprehensively evaluate our model's performance across diverse editing tasks, we construct a benchmark comprising 21 editing tasks. Among them, 11 categories are inherited from and fully overlap with GEdit-Bench~\cite{liu2025step1x}, and 4 logic-related categories are fully sourced from RiseBench~\cite{rise}. The multi-instruction editing task is fully drawn from ComplexEdit~\cite{yang2025texttt}. We additionally introduce 5 new task categories (each with 100 samples) built from data entirely independent of the training set. Following GEdit-Bench, we adopt the Overall Score from VIEScore~\cite{ku2023viescore}, which jointly measures instruction following, subject consistency, naturalness, and artifacts (ranging from 0 to 10) as our evaluation metric. Following \cite{wu2025qwen,liu2025step1x,lin2025uniworldv1,deng2025emerging}, all metrics are evaluated using GPT-4.1.

We also evaluate our method on ImgEdit~\cite{ye2025imgedit}, which encompasses nine representative editing tasks covering diverse editing categories, with a total of 734 real-world test cases. The evaluation metrics include instruction adherence, image editing quality, and detail preservation, each scored from 1 to 5, with all scores assessed by GPT-4.1.
\end{zeroparfillskip}

\noindent\textbf{Training Details}
\begin{zeroparfillskip}
%
%
During the SFT stage, we train 10k steps on 48 GPUs using a 1.5M image–instruction–CoT dataset built from open-source data~\cite{schuhmann2022laion, krasin2017openimages}. Image-instruction pairs are created by (1) instructions generation with Gemini-2.5-Flash under our defined edit taxonomy, (2) image editing with~\cite{labs2025kontext, wu2025qwen, gptimage}, and (3) filtering with both VLM (Gemini-2.5-Flash, GPT-4.1) and human evaluation. The creation of Meta-CoTs follows Section~\ref{subsec:metacotdatapipeline}. The joint 100k understanding data source from LLaVA-OV~\cite{li2024llava} and Mammoth-VL~\cite{guo2024mammoth}.
During the RL stage, we train for 500 steps on an additional 20K editing dataset using 32 GPUs. More details are provided in the supplementary material.
\end{zeroparfillskip}

\begin{table}[t]
\small
\begin{center}
\caption{
\textbf{Comparison of the four components that form the Overall Score in VIEScore across the 21-task benchmark.}
}
\vspace{-5pt}
\setlength{\tabcolsep}{3.68mm}{
\begin{tabular}{l|cccc}
\toprule
\textbf{Method} & 
Ins. &
Con. &
Nat. &
Art. \\
\midrule
Bagel(w/o think) & 6.76& 7.73& 7.01& 7.94\\
Bagel(w think) & 6.30& 8.44& 6.98& 7.71\\
Train Editing Only & 6.61 & 8.22 & 7.18 & 8.06 \\
SFT(\cotname) & 7.23 & 8.53 & 7.26 & 8.25\\
\midrule
\rowcolor{gray!20}SFT + RL (Ours) & 7.44& 8.53& 7.31& 8.34\\
\bottomrule
\end{tabular}}\label{tab:vie_all}
\vspace{-25pt}
\end{center}
\end{table}

\subsection{Quantitative Evaluation}
\begin{zeroparfillskip}
As shown in Table \ref{tab:all_task} and Table \ref{tab:imgedit}, our method achieves notable improvements in the overall editing score, which considers instruction following, consistency, and visual quality. Compared to Bagel (no-think), it achieves +13.1\% on the 21-task benchmark and +19.7\% on ImgEdit. Relative to Bagel (think), gains reach 20.1\% and 13.0\%, respectively.
\end{zeroparfillskip}

To isolate the contribution of the Meta-CoT paradigm, we also train a variant using identical data and optimized with the same parameters, excluding the training of Meta-CoT. As shown in Table \ref{tab:all_task}, our method outperforms this setting by 15.8\%, validating the effectiveness of \cotname. The RL stage also further enhances alignment and stability.

Table \ref{tab:vie_all} provides a breakdown across the four dimensions of VIEScore, Instruction Following, Subject Consistency, Naturalness, and Artifacts, where our method consistently outperforms the baselines, with the largest improvement in Instruction Following. This suggests that Meta-CoT reasoning enhances semantic understanding of both editing operations and targets, leading to more instruction-faithful edits.

\begin{zeroparfillskip}
At the per-task level (as shown in Table \ref{tab:all_task}), our method improves performance on all tasks except text editing. We observe that the reasoning process tends to hinder text editing, likely because the extensive textual reasoning interferes with identifying the correct text to modify. Developing mechanisms to preserve accurate text perception during reasoning remains a promising direction for future work.
\end{zeroparfillskip}

\begin{table}[t]
\small
\begin{center}
\caption{
\begin{zeroparfillskip}
\textbf{Ablation study on (1) the number of meta-tasks defined and tasks trained, and (2) the Task Thinking in \cotname.} \textit{(n meta)} denotes defining n meta-tasks and training only on them. \textit{(5 meta, full-task)} indicates defining 5 meta-tasks, training on full tasks, and decomposing each task's data into meta-tasks.
\end{zeroparfillskip}
}
\vspace{-5pt}
\setlength{\tabcolsep}{3.22mm}{
\begin{tabular}{l|cccc}
\toprule
\textbf{Method} & 
Ins. &
Con. &
Nat. &
Art. \\
\midrule
Train Editing Only & 6.61 & 8.22 & 7.18 & 8.06 \\
\midrule
SFT(3 meta) & 6.75 & 8.33 & 7.15 & 7.86 \\
SFT(4 meta) & 6.93 & 8.44 & 7.17 & 7.94 \\
SFT(5 meta) & 7.09 & 8.48 & 7.20 & 8.10 \\
SFT(6 meta) & 7.13 & 8.51 & 7.22 & 8.07 \\
SFT(5 meta, full-task) & 7.20 & 8.49 & 7.23 & 8.12\\
\midrule
SFT(w/o task think) & 6.98 & 8.35 & 7.19 & 8.07 \\
\midrule
\rowcolor{gray!20}SFT(\cotname) & 7.23 & 8.53 & 7.26 & 8.25\\
\rowcolor{gray!20}SFT + RL(Ours) & 7.44& 8.53& 7.31& 8.34\\
\bottomrule
\end{tabular}}\label{tab:ablation}
\vspace{-20pt}
\end{center}
\end{table}
\begin{table}[ht]
\small
\begin{center}
\caption{\textbf{Ablation study on the amount of visual understanding data mixed during training.} \textit{w/o und.} denotes training without mixing visual understanding data. \textit{CoT} measures the completeness and accuracy of Task Thinking and Target Editing Traversal.}
\vspace{-5pt}
\setlength{\tabcolsep}{2.55mm}{
\begin{tabular}{l|cccc|c}
\toprule
\textbf{Method} & 
Ins. &
Con. &
Nat. &
Art. &
CoT \\
\midrule
Train Editing Only & 6.61 & 8.22 & 7.18 & 8.06 & - \\
\midrule
SFT(w/o und.) & 6.74 & 8.28 & 7.18 & 8.11 & 7.56 \\
SFT(1k und.) & 6.92 & 8.31 & 7.20 & 8.16 & 7.81 \\
\midrule
\rowcolor{gray!20}SFT(\cotname) & 7.23 & 8.53 & 7.26 & 8.25 & 8.89\\
\bottomrule
\end{tabular}}\label{tab:ablation_und}
\vspace{-25pt}
\end{center}
\end{table}
\begin{figure*}[t]
    \centering
    \includegraphics[width=\linewidth]{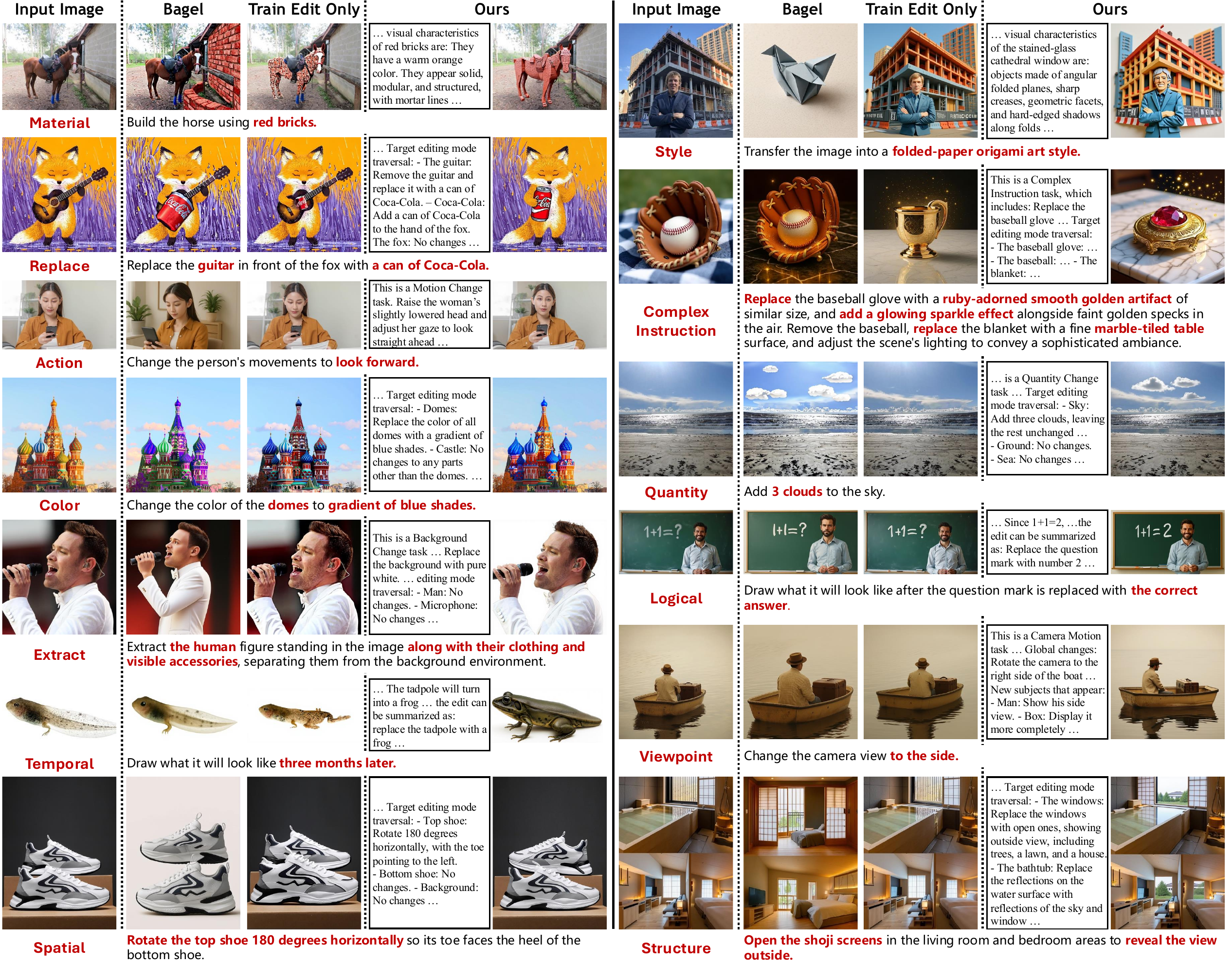}
    \vspace{-20pt}
    \caption{
    \begin{zeroparfillskip}
    \textbf{Qualitative results across diverse editing tasks}, including conventional editing, reasoning-based editing, and multi-instruction editing (Zoom in to view). We present a partial visualization of the \cotname reasoning process. \cotname can decompose instructions, categorize them into specific tasks, generate reasoning based on the task characteristics, and accurately determine whether each target should be edited or not, ultimately achieving better editing results. Please see the supplementary materials for additional task examples.
    \end{zeroparfillskip}
    }
    \label{fig:vis_less}
    \vspace{-10pt}
\end{figure*}

\subsection{Abaltion Study}
We further investigate three critical questions related to our method and present the results in Table \ref{tab:ablation} and Table 
\ref{tab:ablation_und}.

\textbf{Can the meta-task training paradigm enable generalization to unseen editing tasks, and how many meta-tasks are needed?}
As shown in Table \ref{tab:ablation}, we replace the first step of the CoT (``Task Summary") with Meta-task Summary and conduct training under five distinct settings, each corresponding to different definitions of meta-tasks and the number of training task types. Starting from the basic 3-meta-tasks setting (add, delete, replace), we gradually increase the number of meta-tasks.
The results show that, first, the model trained only on the five meta-tasks already achieves performance comparable to the full-data model on the 21-task benchmark and significantly outperforms the train-edit-only version. This demonstrates the strong generalization capability of the meta-task training strategy: training on a small set of meta-tasks while learning task decomposition suffices to generalize to unseen tasks. In other words, mastering universal meta-tasks and task decomposition reasoning enhances the model's ability to generalize.
Second, results show that our defined five meta-tasks strike a good balance between generalization and performance: defining fewer meta-tasks leads to a significant drop in instruction-following, while defining more meta-tasks provides little additional improvement across all tasks.

\textbf{Does the Task Thinking in \cotname benefit the editing process?}
\begin{zeroparfillskip}
As shown in Table \ref{tab:ablation}, we compare our method with a variant that removes the Task Thinking (the second step of Meta-CoT). Results show a significant drop in instruction-following performance, confirming that reasoning based on task characteristics is crucial for the editing.
\end{zeroparfillskip}

\textbf{How does joint training with understanding data affect editing performance?}
%
\begin{zeroparfillskip}
In Table~\ref{tab:ablation_und}, we compare two reduced settings: (a) removing all understanding data and (b) using only 1K samples, against the default 100K. In addition to the four VIEScore components, GPT-4.1 also evaluates CoT quality in terms of the completeness and accuracy of Task Thinking and Target Editing Traversal in Meta-CoT. Results show that both reduced settings cause significant drops in editing performance, particularly in instruction following, as limited understanding data weakens the model's comprehension of editing instructions. This is further corroborated by the notable decline in CoT quality, underscoring the necessity of balancing all three triplet elements during training and demonstrating that higher-quality Meta-CoT reasoning leads to better editing results.
\end{zeroparfillskip}

\subsection{Qualitative Evaluation}
\begin{zeroparfillskip}
As shown in Figure \ref{fig:vis_less}, we present comparisons across diverse editing tasks, including conventional, reasoning-based editing, and multi-instruction editing. Our method significantly improves instruction following, logical reasoning, and multi-instruction understanding compared with baseline methods. This demonstrates that our approach more effectively activates and leverages the model's inherent understanding capability during the editing process.
\end{zeroparfillskip}

\section{Conclusion}
\begin{zeroparfillskip}
In this paper, we have investigated the problem of how to simultaneously enhance the understanding granularity and generalization capability of Chain-of-Thought (CoT)-guided image editing. To address this, we have presented \cotname, which first employs the Triplet Decomposition to stimulate the model's reasoning ability from both task and target perspectives. Furthermore, we have proposed the Meta-task Decomposition, which endows Meta-CoT with strong generalization capability across diverse editing scenarios. To align the CoT reasoning with the editing behavior, we have introduced the \rewardname. Extensive experiments on our proposed 21-task benchmark and ImgEdit have demonstrated that our method not only significantly improves editing performance but also exhibits strong generalization to unseen editing tasks.
\end{zeroparfillskip}

\paragraph{Acknowledgments.} This work was supported part by the Guangdong Natural Science Funds for Distinguished Young Scholar (No. 2025B1515020012).

{
    \small
    \bibliographystyle{ieeenat_fullname}
    \bibliography{main}
}


\end{document}